\documentclass{svproc}

\usepackage{url}

\usepackage{cite}
\usepackage{amsmath,amssymb,amsfonts}
\usepackage{algorithmic}
\usepackage{graphicx}
\usepackage{textcomp}
\usepackage{xcolor}
\usepackage[flushleft]{threeparttable}
\usepackage{tablefootnote}
\usepackage{multirow}
\newcommand{\comment}[1]{}
\PassOptionsToPackage{hyphens}{url}\usepackage{hyperref}
\def\BibTeX{{\rm B\kern-.05em{\sc i\kern-.025em b}\kern-.08em
    T\kern-.1667em\lower.7ex\hbox{E}\kern-.125emX}}
    
\begin{document}
\mainmatter

\title{Deep Reinforcement Learning for Heat Pump Control}
\author{Tobias Rohrer\inst{1} \and Lilli Frison\inst{2}
	Lukas Kaupenjohann\inst{2} \and Katrin Scharf\inst{2} \and Elke Hergenröther\inst{1}}

\authorrunning{Tobias Rohrer et al.} 

\institute{University of Applied Sciences Darmstadt
	\and
	Fraunhofer Institute for Solar Energy Systems}
\maketitle

\begin{abstract}
Heating in private households is a major contributor to the emissions generated today. Heat pumps are a promising alternative for heat generation and are a key technology in achieving our goals of the German energy transformation and to become less dependent on fossil fuels. Today, the majority of heat pumps in the field are controlled by a simple heating curve, which is a naive mapping of the current outdoor temperature to a control action. A more advanced control approach is model predictive control (MPC) which was applied in multiple research works to heat pump control. However, MPC is heavily dependent on the building model, which has several disadvantages. Motivated by this and by recent breakthroughs in the field, this work applies deep reinforcement learning (DRL) to heat pump control in a simulated environment. Through a comparison to MPC, it could be shown that it is possible to apply DRL in a model-free manner to achieve MPC-like performance. This work extends other works which have already applied DRL to building heating operation by performing an in-depth analysis of the learned control strategies and by giving a detailed comparison of the two state-of-the-art control methods.
\end{abstract}

\section{Introduction}

Heating in private households accounted for 26\% of total energy consumed in Germany in 2020,  making it a major contributor to the emissions generated today\cite{heating-stats}. Heat pumps are a sustainable alternative to traditional heating systems which rely on fossil fuels. Heat pumps exploit heat from natural energy sources such as ambient air or groundwater and bring it to a higher temperature level that can be used to supply heat to buildings. Therefore, heat pumps can be used for emission-free room heating if they are driven by electricity generated by renewable sources such as solar or wind. This makes heat pumps one of the key technologies for achieving the goals of the ongoing energy transition \cite{heat_pump_key_tech}.

While the penetration of heat pumps in Germany is steadily increasing \cite{heat-pump-stats}, most heat pumps in the field are controlled by a simple heating curve\cite{control-strategies-heat-pumps}, which is a static mapping from the current outdoor temperature to a control action. While this approach is easy to implement and maintain, it has potential for improvement as factors like a weather forecast or a time varying electricity price do not have any influence on the heating strategy \cite{control-strategies-heat-pumps}.

As a result, model predictive control (MPC) has been applied to heat pump control in the past years. The basic idea of MPC is to make use of a simplified model to predict the effect of control actions. As promising as this sounds, it also introduces a strong dependency on the model which is used. Therefore, the performance of MPC for heat pump control highly depends on the accuracy of the model of the building to be heated \cite{control-strategies-heat-pumps, mpc-beyond-theory, serale2018model}. Additionally, the computational effort required by MPC during execution is relatively high compared to other control methods, which results in additional hardware requibents on the controller level \cite{heat-pumps-smart-grid,  control-strategies-heat-pumps, serale2018model}. This means, the model must be relatively simple in order to achieve reasonable run times, but at the same time it must be accurate enough to make it useful for MPC \cite{serale2018model, privara2013building, wei-deep-rl-hvac}. As a result, often two models must be created in order to apply MPC to heat pump control. One simplified model which can be used by MPC to plan and one more accurate model used to close the control loop during validation\cite{mpc-review, serale2018model}. The process of model creation is complex and considered as the most time-consuming task when applying MPC\cite{serale2018model}. Overall, in recent years, the work on MPC used for heat pump control has increased. However, so far it has not gained acceptance in the field of heat pump control in practice\cite{serale2018model, heat-pumps-smart-grid}. 

Motivated by the shortcomings of MPC and its complex process of model creation, this work applies deep reinforcement learning to heat pump control in a model-free manner. This means, optimal control policies are approximated from experiences collected by interaction with a simulated building, without providing information about the simulation or its underlying model. We use MPC, which utilizes the simulations model as baseline for optimality. This work contributes to the field of research by: (1) formulating heat pump control as continuous control problem, (2) showing through our experiments in a simulated environment that optimal heat pump control strategies can be approximated through a model-free approach by applying deep reinforcement learning, (3) showing that these strategies exploit the thermal storage of the building to shift heating loads to periods where the heat pump can be operated most efficiently and (4) providing a comparison of the two state-of-the-art control methods MPC and DRL.

\section{Related Work}\label{sec:related}
In the last few years, works that applied deep reinforcement learning to heat pump control have increased.The works from \cite{peirelinck-rl-optimize-hp-dr, patyn-rl-hp-nn-architectures, ruelens-residental-demand-response-batch-rl} investigated heat pump control in a demand response setting, where control is supposed to happen with respect to a time varying electricity price signal. They utilized fitted Q-iteration and thus discretized their control actions. With this approach, these works could all report a significant decrease of operating cost\cite{peirelinck-rl-optimize-hp-dr, patyn-rl-hp-nn-architectures, ruelens-residental-demand-response-batch-rl}.

Heidari et al. \cite{heat-pump-warm-water} applied deep Q-learning \cite{atari-dqn} to control a simulated heat pump for hot drinking water usage. While there are also requirements for comfort in this problem statement, their focus can be considered different as their goal was to learn control strategies that are hygiene-aware. They use a binary action space to control the heat pump. By including occupants' warm water usage behaviour, Heidari et al. could enable their agent to learn control strategies that consider occupants' behaviour. They could report savings in energy usage by 24,5\% compared to a rule-based controller which they used as a baseline.

In \cite{heat-pump-heating-net}, Ghane et al. applied deep reinforcement learning by using PPO\cite{schulman2017proximal} to control a simulated heat pump. Their problem statement differs as their goal is to find optimal control strategies for a central heat pump, which provides heat to a heating network consisting of multiple houses. Similar to the work at hand their goal is to minimize electricity usage while meeting the comfort requirements of the occupants. They compared their results against a heating curve based controller. They could report 16.03\% of reduction in energy demand compared to their baseline. They used both discrete and continuous actions to validate their solution. 

In \cite{nagy-deep-rl-optimal-control-space-heating}, Nagy et al. applied deep Q-learning\cite{atari-dqn} to a simulated air-source heat pump. By defining 6 discrete control actions, they managed to learn control strategies that stays in a comfort temperature range while reducing run cost based on an electricity price. They have used a dual price signal with two different fixed prices during the day and night. A baseline comparison using rule based and MPC-based control strategies was conducted. They reported savings of 5-10\% compared to a rule based controller. Like in the work at hand, MPC was used as upper performance limit, as it used the simulation as a model for planning and had thus full knowledge of the target environment. The work from Nagy et al.  \cite{nagy-deep-rl-optimal-control-space-heating} can be considered as the most similar to the work at hand.

We extend other works in the field by implementing heat pump control as continuous control problem. Additionally, we could show by a detailed analysis of the learned control strategies that sophisticated strategies which exploit the thermal storage of the building could be learned.

\section{Problem Formulation}
The main task at hand is to approximate optimal control strategies for an air-source heat pump in simulation in a model-free manner. Thereby, a control strategy must minimize electrical energy usage and comfort deviations at the same time. Comfort deviations occur whenever the indoor room temperature exceeds the comfort temperature range, which is defined between $21^\circ\text{C}$ and $25^\circ\text{C}$. Finding efficient control strategies presents a challenging task, as (1) not only the heat pump operation but also the outside temperature has an impact on the indoor temperature, (2) selected control actions have an effect only with a delay, since the heat transfer does not take place instantaneously, (3) minimization of comfort deviations and electrical usage are conflicting objectives and (4) the outdoor temperature effects the efficiency of the heat pump as it uses the outdoor air as heat source\footnote{The lower the difference between the heat source and the target temperature, the higher the efficiency of a heat pump.}. 


\subsection{Simulation Framework}\label{sec:sim}
\begin{figure}
	\centering
	\includegraphics[width=120mm]{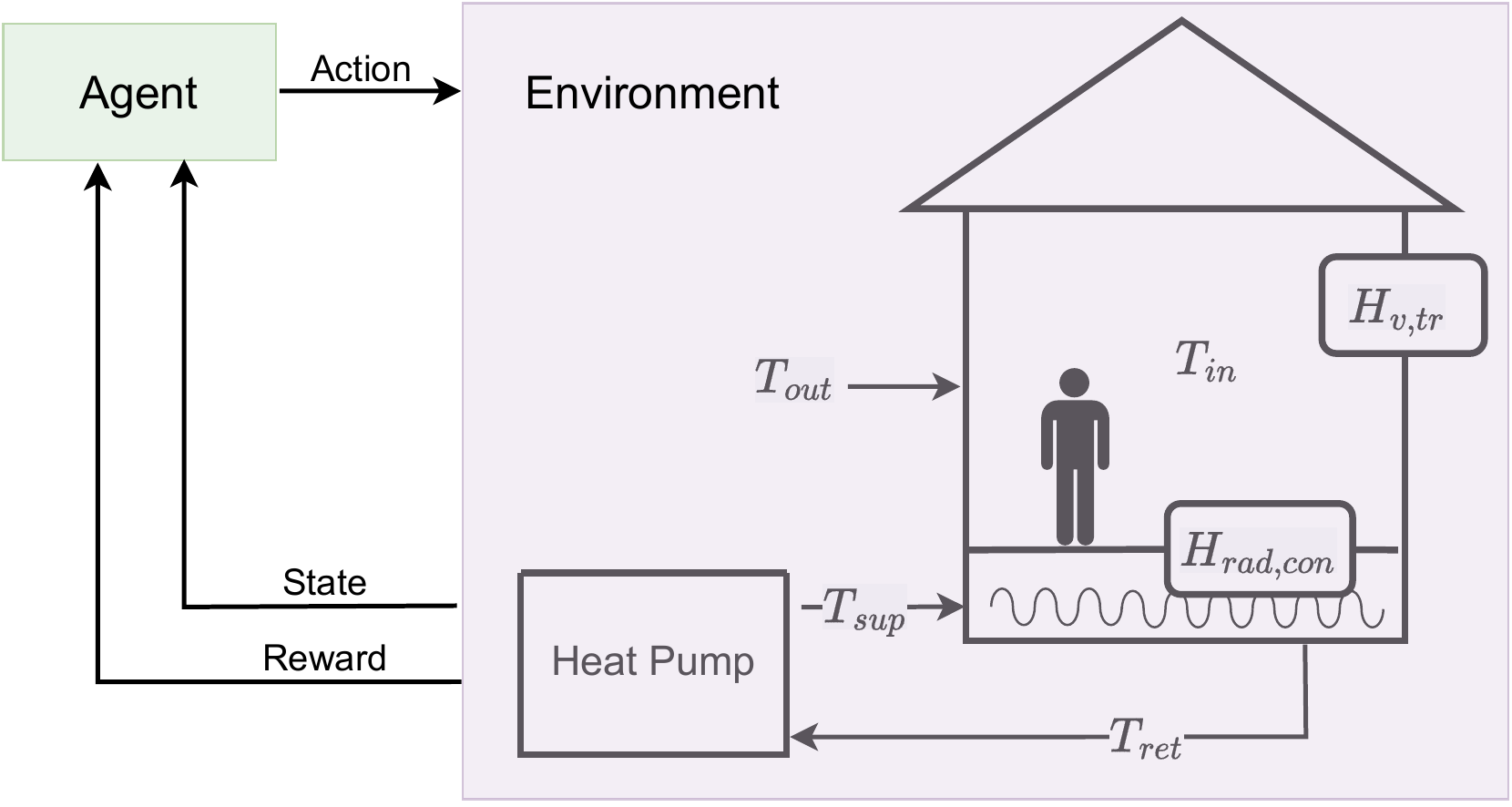}
	\caption[Overview of the Simulation Framework]{Overview of the simulation framework and the interaction between agent and environment.}
	\label{fig:simulation}
\end{figure}

The core functionality of the simulation framework used in this work is illustrated by Figure \ref{fig:simulation}. It mimics a simplified building with a single room, which is heated by a floor heating system supplied with heat which is generated by an air source heat pump. The framework simulates the effect of the outdoor temperature $T_{out}$ and heat pump supply temperature $T_{sup}$ on the building indoor temperature $T_{in}$. The supply temperature defines the temperature of the water heated by the heat pump which enters the floor heating system. Additionally, the simulation calculates the return temperature $T_{ret}$ which describes the temperature of the water coming back from the floor heating system which is to be heated again by the heat pump. The heat transfer between outdoor and indoor temperature is parameterized by the coefficient for transmission and ventilation $H_{ve,tr}$. Similarly, the heat transfer coefficient for radiation and convection $H_{rad,con}$ is used to model the heat transfer between indoor temperature and the floor heating system. One simulation step is modelled as if 900 seconds would pass in real-time.

At its heart, the simulation framework utilizes a 2R2C lumped capacitance model which can be expressed by the following differential equations:
\begin{equation}
    \label{eq:2R2C_dgl1}
    \frac{dT_{in}}{dt} = \frac{1}{C_{bldg}} \cdot (H_{rad,con} \cdot (T_{ret} – T_{in}) – H_{ve,tr} \cdot (T_{in} – T_{out}))
\end{equation}
\begin{equation}
    \label{eq:2R2C_dgl2}
    \frac{dT_{ret}}{dt} = \frac{1}{C_{water}} \cdot (\dot{Q}_{hp} - H_{rad,con} \cdot (T_{ret} – T_{in}))
\end{equation}
Hereby, the term $C_{bldg}$ defines the thermal capacity of the building envelope and $C_{water}$ defines the thermal capacity of the water in the underfloor heating system. $\dot{Q}_{hp}$ defines the thermal power generated by the heat pump during a time step and is the variable which needs to be controlled by the control strategy.

The framework can be used to simulate a wide variety of buildings by setting different parameters like the transmission losses and heat capacities. The parameters and buildings used in this work to evaluate the proposed solution will be explained in more detail in Section \ref{sec:experiments}.

\section{Deep Reinforcement Learning for Heat Pump Control}
We applied deep reinforcement learning to approximate optimal heat pump control strategies. Deep reinforcement learning targets the learning of strategies through interactions between an \emph{agent} and an \emph{environment}. Thereby, the agents goal is to choose \emph{actions} which maximize the sum of \emph{rewards} over time. The reward serves as feedback on the quality of a certain action in a given \emph{state} \cite{rl_intro_book}. This interaction between agent and environment is illustrated in Figure \ref{fig:simulation} and will be explained in more detail in Sections \ref{sec:mdp} and \ref{sec:rl}.

\subsection{Heat Pump Control as Markov Decision Process}\label{sec:mdp}
The Markov decision process (MDP) serves as a mathematical formalism of sequential decision making problems. It is used to formally define environments in reinforcement learning \cite{rl_intro_book}. The following describes the components that define the MDP used this work:

\subsubsection{Action Space}
The action which can be chosen by the agent represents the thermal power $\dot{Q}_{hp}$ which is generated by the heat pump during the upcoming time step. The action space is continuous in the range $\dot{Q}_{hp} \in [0,12kW]$. The action space is internally rescaled, so the actions taken by the agent lie in the symmetric interval $[-1,1]$. 

\subsubsection{State Space} 
A state is defined as $\{T_{in}^t, T_{ret}^t, T_{out}^t, T_{out}^{t+1}, T_{out}^{t+2}, ..., T_{out}^{t+n}\}$, where $T_{in}^t$ defines the indoor temperature, $T_{ret}^t$ the return temperature, and  $T_{out}^t$ the outdoor air temperature of the current time step $t$. Additionally, the state contains a perfect forecast of the outdoor temperature of the next $n$ time steps $\{T_{out}^{t+1}, T_{out}^{t+2}, ..., T_{out}^{t+n}\}$. As shown later, the number of forecast steps included depends on the building at hand. It is important to note that the features contained in the state are available in different orders of magnitude. Therefore, standardization of the states was done by applying moving average standardization as implemented in \cite{stable-baselines3}. 


\subsubsection{Reward Function}
The reward encodes the goal of the research task at hand. Therefore, it balances the minimization of electricity usage and comfort deviations. The reward is calculated at every time step according to the following formula:
\begin{equation}\label{eq:reward}
r_t  = -1 * (\beta * electricity\_used_t + comfort\_deviation_t).
\end{equation}
The trade-off parameter $\beta$ is used to balance the conflicting objectives of minimizing comfort deviations and electrical usage.

\subsection{Deep Reinforcement Learning Agent}\label{sec:rl}
The agent is represented by a policy $\pi$ which we implemented as fully connected neural network with two hidden layers containing 64 neurons each. In the following we refer to this network as \emph{policy network}. During training, the policy network learns to approximate optimal heat pump control actions based on a state. Out of the many deep reinforcement learning algorithms that exist, we chose Proximal Policy Optimization (PPO) \cite{schulman2017proximal} to train the agent. We chose PPO as it (1) supports continuous action spaces, (2) was applied lately to problems which can be considered more complex \cite{openai-dota, openai2018learning}, and (3) was reported as a default reinforcement learning algorithm at OpenAI, one of the pioneers in the field of deep reinforcement learning \cite{ppo_openAI}. We utilized the deep reinforcement learning framework Stable Baselines3 \cite{stable-baselines3} in order to implement and train the agent.

\section{Experiment Setup}\label{sec:experiments}
\begin{table}[]
	\centering
	\small
		\caption{Summary of Simulated Buildings}
		\label{tab:buildings}
		\begin{tabular}{lllll}
			\cline{1-4}
			& Old & Efficient &  \\ \cline{1-4}
			Floor Size in $m^2$ & 136 & 393  &  \\
			Heat Capacity $C_{bldg}$ in $Wh/m^2/K$ & 45 & 65.9 &  \\
			Transmission Losses $H_{ve,tr}$ in $W/K$ & 396 & 281.7 &  \\
			Energy-Efficiency Class\tablefootnote{According to the German buidling energy act \url{https://www.bmwsb.bund.de/Webs/BMWSB/DE/themen/bauen/energieeffizientes-bauen-sanieren/gebaeudeenergiegesetz/gebaeudeenergiegesetz-artikel.html} } & F & A& \\
			Year of Construction &1984 & 2020 & \\
			\cline{1-4}
		\end{tabular}
\end{table}



We used the simulation framework described in section \ref{sec:sim} to imitate two different buildings. An \emph{old} building, which can be considered inefficient in terms of its thermal properties and an \emph{efficient} building, which can be considered energy efficient. See Table \ref{tab:buildings} for a more detailed description of the buildings. The heat pump simulated corresponds to a Dimplex LA 6TU air source heat pump with maximum heating power set to 12kW.

We used weather data from the photovoltaic geographical information system of the European Commission\footnote{\url{https://re.jrc.ec.europa.eu/}} to feed the simulation with outdoor air temperature data. We used the weather data between 2010 and 2015 for training and left the data from 2016 untouched for obtaining results. Weather data from April to September were excluded in both cases as heating in those months is usually not necessary. In the following, we are referring to the data from 2010 to 2015 as \emph{training data} and to the data from 2016 as \emph{testing data}. 

\comment{The data contains information about the weather in Freiburg, Germany between 2010 and 2016.}

We trained one agent per building independently, which led to two independent agents. The training was conducted over 350 episodes, each containing 2880 interactions between agent and environment. Therefore, each of the two agents was trained over 1,000,000 time steps. One training episode corresponds to approximately one month, as one simulation step represents 900 seconds in real time. One month of weather data was chosen at random from the training data at the beginning of each episode. 
\comment{After every seventh episode of training, each of the agents was \emph{validated}. This included pausing the training by pausing exploration and updates to the neural network and running the agent for one episode. By doing so we could collect data about the agents performance without the randomness introduced by exploration. This was necessary to get the best performing agent which occurred during training as performance improvement is not guaranteed to happen monotonically.}
We repeated the training for both of the agents five times using different preset random seeds, as training might run significantly different for different seeds\cite{rl-reproducing}. After training, we chose the agent which performed best out of the five runs and discarded the others.

We setup MPC to serve as baseline for optimality by using the simulation framework itself to plan and execute the control actions. This would mostly not be applicable in reality, as usually, a simplified model is used to plan and a more complex model or the reality is used to close the control loop by executing the planned control actions \cite{mpc-review, serale2018model}. However, in our scenario, providing MPC the same model for planning and execution gives us an optimal baseline for comparison.

\comment{
\begin{figure}
	\centering
	\includegraphics[width=85mm]{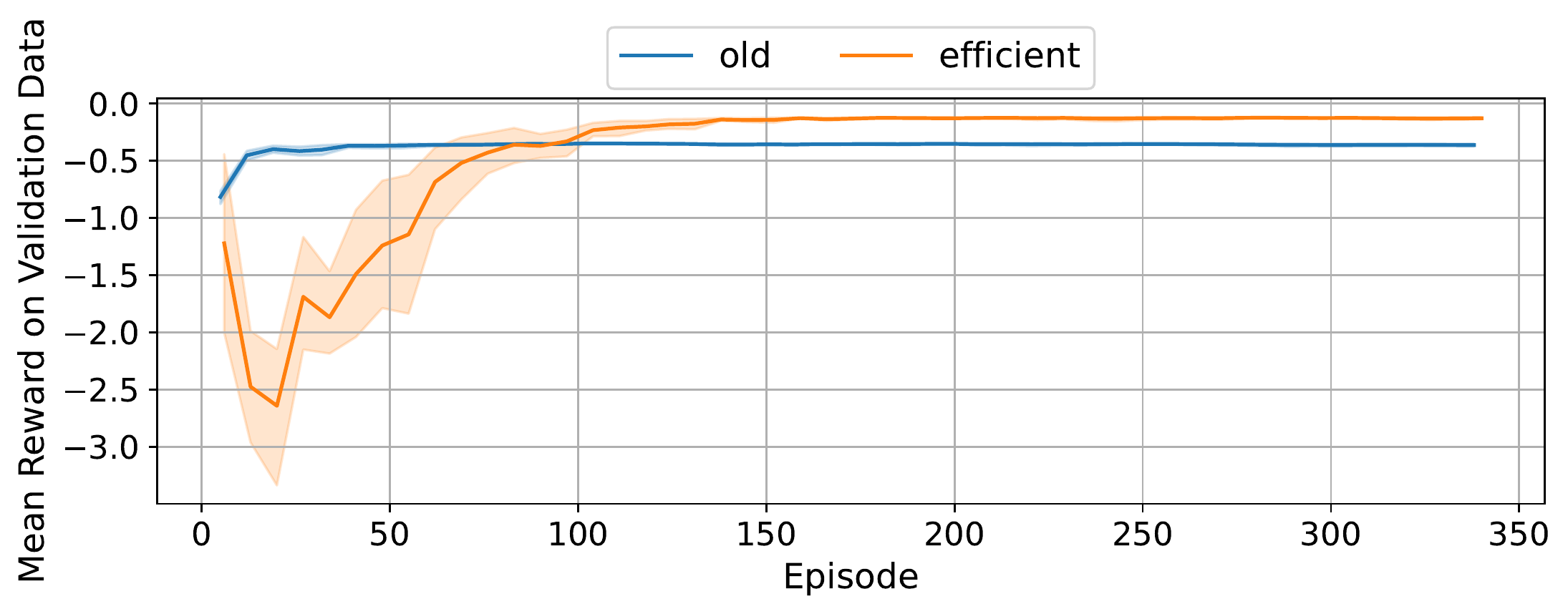}
	\caption[Evolution of Mean Reward]{Evolution of the mean reward on the validation data per time step during training for each of the three buildings. For each building, the training was executed five times using different preset random seeds. The shaded areas show the variance of the progression of the mean reward caused by the different preset random seeds. The line indicates the mean progression.}
	\label{fig:train_progress}
\end{figure}}

\section{Results}
To obtain the results, the agents were executed using weather profiles of the six months from 2016 which were not used during training and therefore represent our testing data.
\subsection{Qualitative Results}
Figure \ref{fig:strategies} shows the control strategies picked up by the agents during training. It can be observed, that the learned control strategies differ between both buildings. 
\subsubsection{Old Building}
\begin{figure}
	\centering
	\includegraphics[width=125mm]{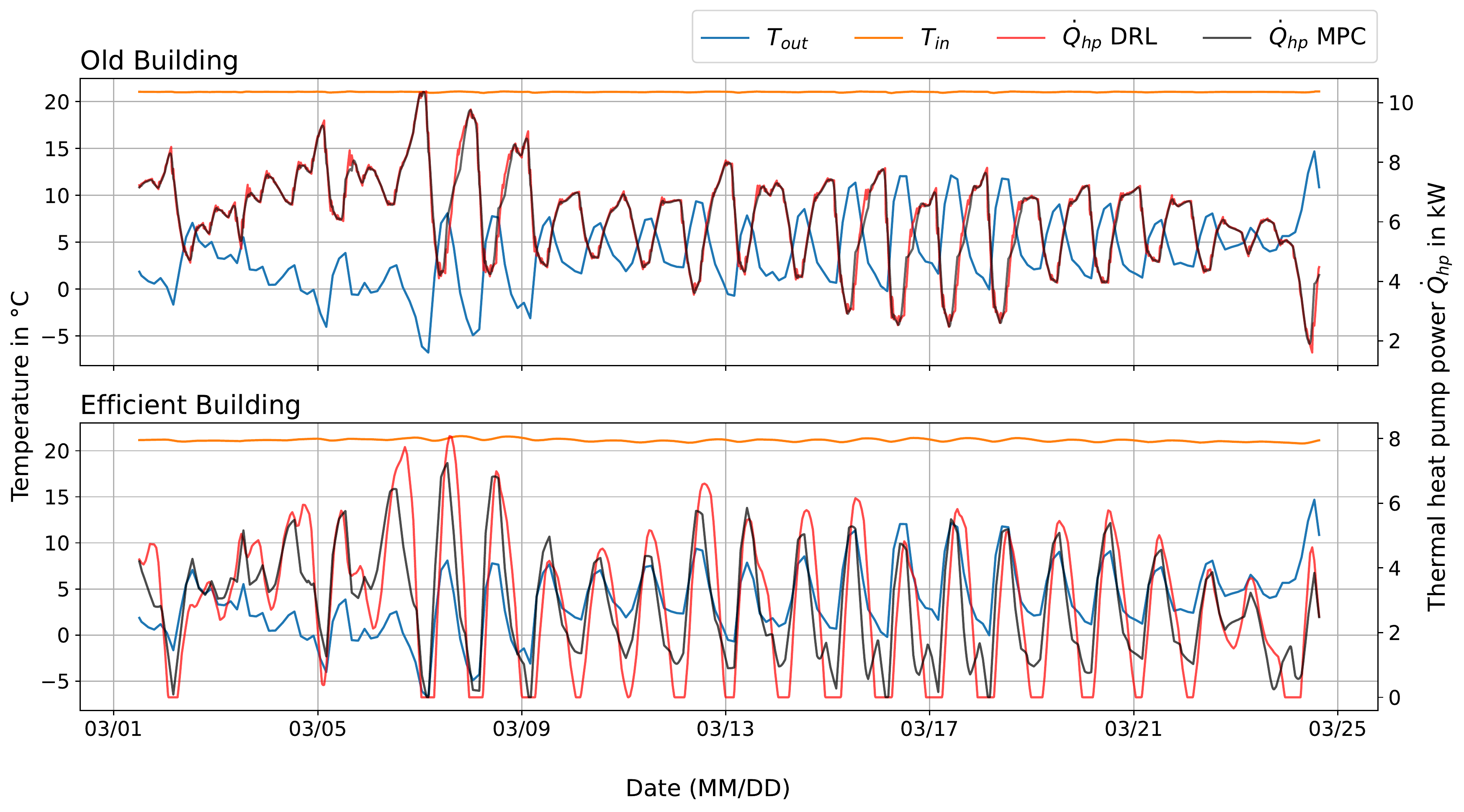}
	\caption{Comparison of strategies taken by the different control methods when heating the different. In both cases, the same weather profile from March from the testing data was used. Plots of control during other months of the testing dataset can be found in \cite{me-master}}
	\label{fig:strategies}
\end{figure}

As shown in the top part of Fig. \ref{fig:strategies}, the trained agent of the old  building regulates the indoor temperature almost constantly at $21^\circ\text{C}$. Intuitively, this strategy is reasonable, as $21^\circ\text{C}$ marks the lower comfort bound. Heating further would not increase comfort by our definition, but lead to an increase in electricity usage. Experiments have shown that the agent does not benefit from weather forecasts that go far into the future. A prediction length of 8 time steps which corresponds to two hours performed best. Longer forecasts worsened performance due to an increase in the dimensionality of the observation space. Thus, we interpret the agent learned a strategy which can be considered relatively shortsighted. Which makes sense, as storing heat in case of the old building comes with high losses. The efficiency of this strategy is backed up as MPC, which is our baseline for optimality, takes a strategy which is almost identical (see top Fig. \ref{fig:strategies}).
 
\comment{Analogously, a relatively low $\gamma$ of 0.96 performed best, as it minimizes the impact of time distant rewards.}

\subsubsection{Efficient Building}
In contrast to the old building, the agent learned a strategy where the majority of heating is taking place while the outside temperature is relatively high (see bottom part of Fig. \ref{fig:strategies}). This makes sense as the efficiency of the heat pump increases with the temperature of the low-temperature source. The agent learned to exploit the higher heat capacity and lower transmission losses of the efficient building (shown in Table \ref{tab:buildings}) by storing heat and thus shift the heating load to periods, where the heat pump can be operated more efficiently. For this strategy, the agent learned to utilize the weather forecast. Experiments showed that a temperature forecast of 48 time steps which corresponds to 12 hours performed best.\comment{Analogously, a relatively high $\gamma$ of 0.99 was chosen, as actions at a current time step have a relatively high impact on future rewards.} When comparing the learned strategy to MPC, we can see that both stategic approaches are quite similar.

\subsection{Quantitative Results}

Table \ref{tab:baseline_compare_} quantifies the results of the strategies discussed above. It can be seen that DRL consumes only marginally (less than 1\%) more energy for both buildings compared to MPC, which as explained above can be considered as baseline for optimality in our setup. The execution times indicate the time required by MPC and the DRL agent to perform heat pump control during the six test episodes. This includes 17,280 simulation steps and the same amount of control decisions to be taken. Here it can be seen that DRL is many times faster during execution\footnote{During execution, the DRL agent approximates the optimal control action by forward propagating the neural network through fast matrix operations. MPC has to solve an optimization problem at every time step which comes with cost of compute.}.
\begin{table}[]
	\centering
	\small
	\begin{threeparttable}
		\caption{Quantitative Baseline Comparison}
		\label{tab:baseline_compare_}
		\begin{tabular}{llll}\cline{1-4} 
			&  & DRL & MPC \\ \cline{1-4} 
			\multirow{3}{*}{Old} 
			& Electricity Mean in Wh & 405.15 & 403.23\\
			& Comfort Deviation Mean in $^\circ\text{C}$ & 0 & 0.02 \\
			& Comfort Deviation Max in $^\circ\text{C}$& 0.18 & 0.15 \\ 
			\cline{1-4} 
			\multirow{3}{*}{Efficient} 
			& Electricity Mean in Wh & 137.92 & 137.65 \\
			& Comfort Deviation Mean in $^\circ\text{C}$& 0.02 & 0 \\
			& Comfort Deviation Max in $^\circ\text{C}$& 0.21 & 0.20 \\ 
			\cline{1-4} 
			& Execution Time in Seconds& 38 & 1679 \\ \cline{1-4} 
		\end{tabular}
		\begin{tablenotes}
			\small
			\item \textbf{Note:} Values were rounded to two decimal places. The mean values were calculated over the whole testing data and are to be considered per time step, which simulates 900s in real time. The Max denotes the maximum comfort deviation which occurred any time during the control period of the whole testing data.
		\end{tablenotes}
	\end{threeparttable}
\end{table}

\section{Extension to a Demand Response Scenario}
The purpose of this experiment was to show if the proposed method can be extended to a demand response scenario, where control should happen with respect to a time varying electricity price signal.

\subsection{Setup Demand Response Scenario}
To the best of our knowledge, there is no publicly available data about time varying electricity prices for residential customers in Europe. Therefore, like in \cite{sweetin-master}, we used day-ahead electricity prices from the European power exchange (EPEX) spot marked. These prices are used to balance supply and demand between electricity producers and distributors. It must be noted, that the used EPEX spot prices differ from residential customer electricity prices as they do not include taxes and costs for grid usage. The price data used in this experiment originate from the same time period as the weather data and were provided by the Fraunhofer Institute for Solar Energy Systems. Like the weather data, the price data was also split into training and testing data. In order to perform the demand response experiment, we changed the MDP definition as follows:
\subsubsection{Reward}
The new objective of the agent is to minimize the \emph{operational cost} of the heat pump while still maintaining comfort. The operational cost result from the amount of energy consumed and the electricity price at that time. This new objective was encoded in the reward as follows:

\small
\begin{equation}\label{eq:reward_price}
r_t  = -1 * (\beta *electricity\_used_t * price_t + comfort\_deviation_t)
\end{equation}
\normalsize

Note that the new reward definition is analogue to the one from \eqref{eq:reward}, but differs as the price is included. Analogous, a trade-off parameter $\beta$ is used to balance comfort and costs.
\subsubsection{State} We included a perfect forecast of the electricity price in the state to give the agent the possibility to decide depending on the price. This lead to the following state definition:  $\{T_{in}^t, T_{ret}^t, T_{out}^t, price^{t}, T_{out}^{t+1}, price^{t+1}, T_{out}^{t+2}, price^{t+2}, ...\}$. A forecast length of 32 time steps, which corresponds to 8 hours of the price and outside temperature has shown to perform best.

We trained the agent analogously to the procedure described in Section \ref{sec:experiments}. The experiment was conducted with the efficient building only, as the old building did not provide any heat storage capabilities to shift heating from high to low price periods.
\subsection{Results Demand Response Scenario}
\begin{table}[]
	\centering
	\small
	\begin{threeparttable}
		\caption{Results of the Demand Response Scenario}
		\label{tab:dr}
		\begin{tabular}{llll}\cline{1-4} 
			& DRL & MPC & DRL Baseline \\ \cline{1-4} 
			Cost Mean in € Cent & 0.32 & 0.31  &  0.47 \\
			Comfort Deviation Mean $^\circ\text{C}$ & 0.2 & 0.22  & 0.02 \\
			Comfort Deviation Max $^\circ\text{C}$& 1.08 & 0.85  & 0.21 \\ \cline{1-4}
			Execution Time in Seconds& 38 & 1677 & 38 \\ \cline{1-4}  
		\end{tabular}
		\begin{tablenotes}
			\small
			\item \textbf{Notes:} Values were rounded to two decimal places and the statistics are calculated over the whole testing data. The description of Table \ref{tab:baseline_compare_} provides more details on how the mean and Max were calculated.
		\end{tablenotes}
	\end{threeparttable}
\end{table}

\begin{figure*}
	\centering
	\includegraphics[width=125mm]{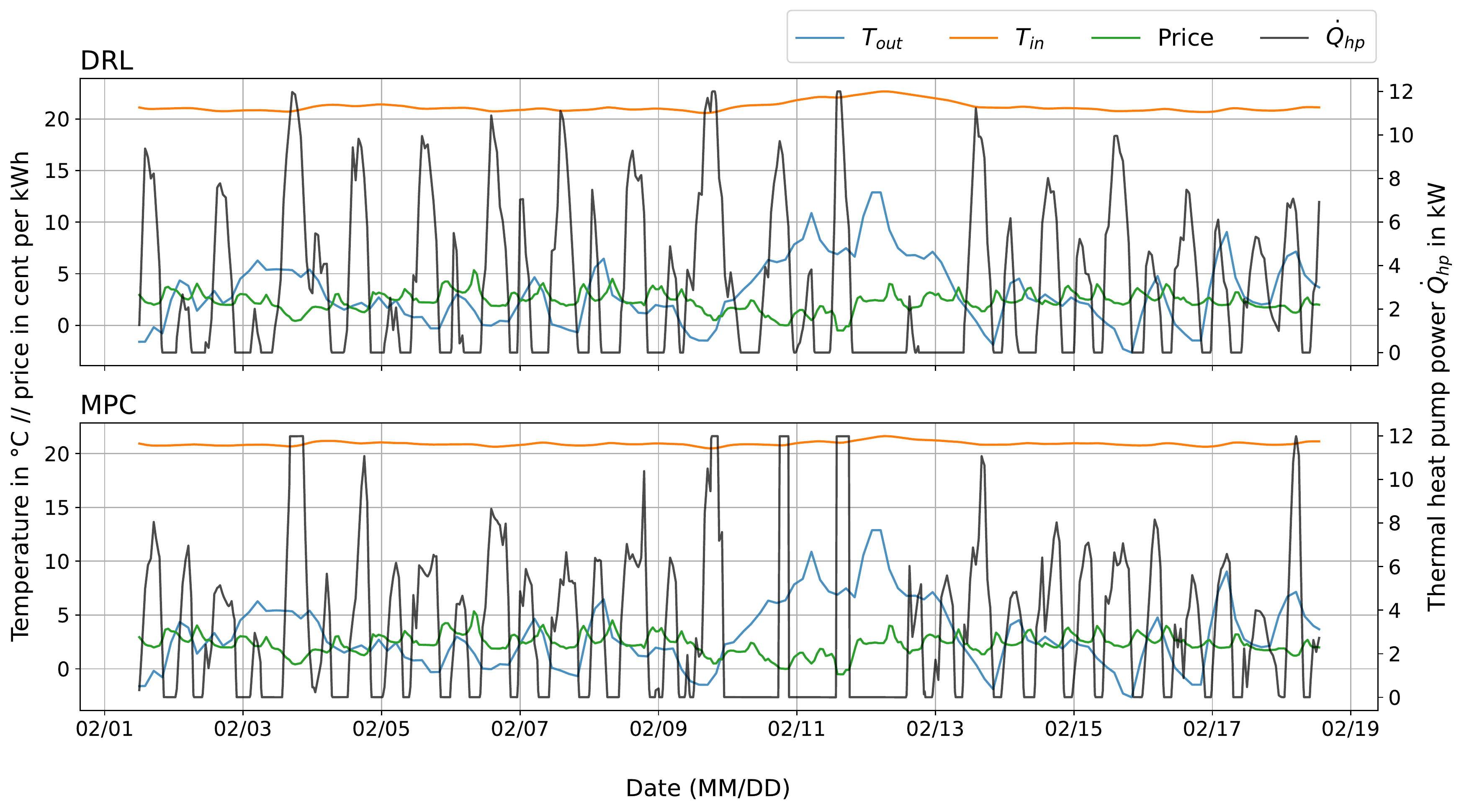}
	\caption{Heat pump control in a demand response scenario by MPC and the deep reinforcement learning agent for the efficient building. For both control methods, the same weather profile from February from the testing data was used. Plots of control during other months of the testing data can be found in \ref{me-master}}
	\label{fig:example_control_dr}
\end{figure*}

Figure \ref{fig:example_control_dr} demonstrates the functionality of the trained agent for the efficient building in the demand response scenario. It can be seen that the agent learned to shift the heating to low price periods. Table \ref{tab:dr} quantifies the results of this experiment. The cost and comfort deviation mean values denote the mean per time step and are calculated using all months contained in the testing data. Based on the costs, it can be seen that the proposed solution is almost as effective as MPC in the demand response context, which can be considered as the optimal solution. Additionally, the DRL method from Section \ref{sec:experiments}, which controls independently of the price signal is listed in the Table and serves as an additional baseline. Although the DRL method described in this section causes more comfort deviations, operational costs could be reduced by almost a third compared to the DRL baseline. It must be noted, that the quantification of the cost saving potential depends on the price signal used and must be therefore interpreted with caution.

\section{Conclusion and Future Work}

Motivated by the shortcomings of the traditional heating curve and MPC, this work presents an model-free approach to heat pump control by applying deep reinforcement learning. The results indicate, that it is possible to learn optimal heat pump control policies just by learning from interactions with the simulated building in a model-free manner. However, our work and the related works listed in Section \ref{sec:related} were applied in simulation and not to the real world. Therefore, in future works we will focus on the transition of the presented concepts from simulation to reality. 

\newpage
\bibliographystyle{ieeetr}
\bibliography{bibbl.bib}

\begin{thebibliography}{10}

\bibitem{heating-stats}
{AG Energiebilanzen e.V.}, ``{Anwendungsbilanzen zur Energiebilanz
  Deutschland}.''
  \textsc{url:}~\url{https://ag-energiebilanzen.de/wp-content/uploads/2020/10/ageb_20v_v1.pdf}.
\newblock (accessed: Apr. 2, 2022).

\bibitem{heat_pump_key_tech}
P.~Sterchele, J.~Brandes, J.~Heilig, D.~Wrede, C.~Kost, T.~Schlegl, A.~Bett,
  and H.~Henning, ``{Wege zu einem klimaneutralen Energiesystem. Die deutsche
  Energiewende im Kontext gesellschaftlicher Verhaltensweisen}.'' Fraunhofer
  ISE, 2021.
\newblock (accessed: May 23, 2022).

\bibitem{heat-pump-stats}
{Bundesverband Wärmepumpe e.V.}
  \textsc{url:}~\url{https://www.waermepumpe.de/presse/zahlen-daten/}.
\newblock (accessed: Apr. 21, 2022).

\bibitem{control-strategies-heat-pumps}
D.~Rolando and M.~Hatef, ``Smart control strategies for heat pump systems.''
  KTH Royal Institute of Technology,
  \textsc{url:}~\url{https://varmtochkallt.se/wp-content/uploads/Projekt/EffsysExpand/P18_Project_Report_final_reviewed.pdf},
  2018.
\newblock (accessed: Mar. 14, 2022).

\bibitem{mpc-beyond-theory}
J.~C{\'\i}gler, D.~Gyalistras, J.~{\v{S}}iroky, V.~Tiet, and L.~Ferkl, ``Beyond
  theory: the challenge of implementing model predictive control in
  buildings,'' in {\em Proceedings of 11th Rehva world congress, Clima},
  vol.~250, 2013.

\bibitem{serale2018model}
G.~Serale, M.~Fiorentini, A.~Capozzoli, D.~Bernardini, and A.~Bemporad, ``Model
  predictive control (mpc) for enhancing building and hvac system energy
  efficiency: Problem formulation, applications and opportunities,'' {\em
  Energies}, vol.~11, no.~3, 2018.

\bibitem{heat-pumps-smart-grid}
D.~Fischer and H.~Madani, ``On heat pumps in smart grids: A review,'' {\em
  Renewable and Sustainable Energy Reviews}, vol.~70, pp.~342--357, 2017.

\bibitem{privara2013building}
S.~Privara, J.~Cigler, Z.~V{\'a}{\v{n}}a, F.~Oldewurtel, C.~Sagerschnig, and
  E.~{\v{Z}}{\'a}{\v{c}}ekov{\'a}, ``Building modeling as a crucial part for
  building predictive control,'' {\em Energy and Buildings}, vol.~56,
  pp.~8--22, 2013.

\bibitem{wei-deep-rl-hvac}
T.~Wei, Y.~Wang, and Q.~Zhu, ``Deep reinforcement learning for building hvac
  control,'' in {\em Proceedings of the 54th annual design automation
  conference 2017}, pp.~1--6, 2017.

\bibitem{mpc-review}
A.~Afram and F.~Janabi-Sharifi, ``Theory and applications of hvac control
  systems – a review of model predictive control (mpc),'' {\em Building and
  Environment}, vol.~72, pp.~343--355, 2014.

\bibitem{peirelinck-rl-optimize-hp-dr}
T.~Peirelinck, F.~Ruelens, and G.~Decnoninck, ``Using reinforcement learning
  for optimizing heat pump control in a building model in modelica,'' in {\em
  2018 IEEE International Energy Conference (ENERGYCON)}, pp.~1--6, 2018.

\bibitem{patyn-rl-hp-nn-architectures}
C.~Patyn, F.~Ruelens, and G.~Deconinck, ``Comparing neural architectures for
  demand response through model-free reinforcement learning for heat pump
  control,'' in {\em 2018 IEEE International Energy Conference (ENERGYCON)},
  pp.~1--6, 2018.

\bibitem{ruelens-residental-demand-response-batch-rl}
F.~Ruelens, B.~Claessens, S.~Vandael, B.~De~Schutter, R.~Babuska, and
  R.~Belmans, ``Residential demand response applications using batch
  reinforcement learning,'' {\em arXiv preprint arXiv:1504.02125}, 2015.

\bibitem{heat-pump-warm-water}
A.~Heidari, F.~Marechal, and D.~Khovalyg, ``An adaptive control framework based
  on reinforcement learning to balance energy, comfort and hygiene in heat pump
  water heating systems,'' {\em Journal of Physics: Conference Series},
  vol.~2042, p.~012006, nov 2021.

\bibitem{atari-dqn}
V.~Mnih, K.~Kavukcuoglu, D.~Silver, A.~A. Rusu, J.~Veness, M.~G. Bellemare,
  A.~Graves, M.~Riedmiller, A.~K. Fidjeland, G.~Ostrovski, S.~Petersen,
  C.~Beattie, A.~Sadik, I.~Antonoglou, H.~King, D.~Kumaran, D.~Wierstra,
  S.~Legg, and D.~Hassabis, ``Human-level control through deep reinforcement
  learning,'' {\em Nature}, vol.~518, no.~7540, pp.~529--533, 2015.

\bibitem{heat-pump-heating-net}
S.~Ghane, S.~Jacobs, W.~Casteels, C.~Brembilla, S.~Mercelis, S.~Latré,
  I.~Verhaert, and P.~Hellinckx, ``Supply temperature control of a heating
  network with reinforcement learning,'' in {\em 2021 IEEE International Smart
  Cities Conference (ISC2)}, pp.~1--7, 2021.

\bibitem{schulman2017proximal}
J.~Schulman, F.~Wolski, P.~Dhariwal, A.~Radford, and O.~Klimov, ``Proximal
  policy optimization algorithms,'' {\em arXiv preprint arXiv:1707.06347},
  2017.

\bibitem{nagy-deep-rl-optimal-control-space-heating}
A.~Nagy, H.~Kazmi, F.~Cheaib, and J.~Driesen, ``Deep reinforcement learning for
  optimal control of space heating,'' {\em arXiv preprint arXiv:1805.03777},
  2018.

\bibitem{rl_intro_book}
R.~S. Sutton and A.~G. Barto, {\em Reinforcement Learning: An Introduction}.
\newblock MIT press, second~ed., 2018.

\bibitem{stable-baselines3}
A.~Raffin, A.~Hill, A.~Gleave, A.~Kanervisto, M.~Ernestus, and N.~Dormann,
  ``Stable-baselines3: Reliable reinforcement learning implementations,'' {\em
  Journal of Machine Learning Research}, vol.~22, no.~268, pp.~1--8, 2021.

\bibitem{openai-dota}
{OpenAI et al.}, ``Dota 2 with large scale deep reinforcement learning,'' {\em
  arXiv preprint arXiv:1912.06680}, 2019.

\bibitem{openai2018learning}
OpenAI, M.~Andrychowicz, B.~Baker, M.~Chociej, R.~Józefowicz, B.~McGrew,
  J.~Pachocki, A.~Petron, M.~Plappert, G.~Powell, A.~Ray, J.~Schneider,
  S.~Sidor, J.~Tobin, P.~Welinder, L.~Weng, and W.~Zaremba, ``Learning
  dexterous in-hand manipulation,'' {\em CoRR}, 2018.

\bibitem{ppo_openAI}
OpenAI, ``Proximal policy optimization.''
  \textsc{url:}~\url{https://openai.com/blog/openai-baselines-ppo/}.
\newblock (accessed: Sep 10, 2022).

\bibitem{rl-reproducing}
P.~Henderson, R.~Islam, P.~Bachman, J.~Pineau, D.~Precup, and D.~Meger, ``Deep
  reinforcement learning that matters,'' in {\em Proceedings of the AAAI
  conference on artificial intelligence}, vol.~32, 2018.

\bibitem{me-master}
T.~Rohrer, ``Deep reinforcement learning for heat pump control,'' Master's
  thesis, Darmstadt University of Applied Sciences, 2022.

\bibitem{sweetin-master}
S.~Paul, ``Learning to control heat pumps using supervised and reinforcement
  learning methods based on neural networks,'' Master's thesis,
  Albert-Ludwigs-University Freiburg, 2019.

\end{thebibliography}

\end{document}